\documentclass[numbered,webpdf,modern,medium]{oup-authoring-template}
\usepackage{lmodern}
\usepackage[T1]{fontenc}

\usepackage{endnotes}
\let\footnote\endnote
\graphicspath{{Fig/}}
\newcommand{\societylogo}{}
\renewcommand{\figurename}{Fig.}
\makeatletter
\long\def\@figurecaption#1#2{\vskip3pt%
  \begingroup\centering\fontsize{8bp}{11}\selectfont%
  \textbf{#1}\space#2\strut\par\endgroup}
\long\def\@tablecaption#1#2{%
  \begingroup\centering\fontsize{9bp}{11.5pt}\sffamily\selectfont%
  \textbf{#1}\space{#2\strut\par}\endgroup}
\makeatother

\makeatletter
\renewcommand\@seccntformat[1]{%
  \ifnum\pdfstrcmp{#1}{section}=0 \csname the#1\endcsname.\hskip 0.5em
  \else \csname the#1\endcsname\hskip 0.5em\fi}
\makeatother

\makeatletter
\def\ps@opening{\let\@oddhead\@empty\let\@evenhead\@empty\let\@oddfoot\@empty\let\@evenfoot\@empty}
\def\ps@headings{%
  \def\@oddhead{\hfill\sffamily\fontsize{7bp}{9bp}\selectfont\thepage}%
  \let\@evenhead\@oddhead
  \let\@oddfoot\@empty\let\@evenfoot\@empty}
\makeatother

\begin{document}

\journaltitle{Digital Scholarship in the Humanities}
\copyrightyear{2026}
\pubyear{2026}

\firstpage{1}

\title[Crossing the Gender Binary]{Crossing the Gender Binary: Modelling Gendered Reference in Science Fiction and General Fiction}

\author[1,$\ast$]{Yuxuan Liu}
\author[2]{Haim Dubossarsky}
\author[3]{Ruth Ahnert}

\address[1]{\orgdiv{School of the Arts}, \orgname{Queen Mary University of London}, \orgaddress{\street{Mile End Road}, \postcode{E1 4NS}, \state{London}, \country{United Kingdom}}}
\address[2]{\orgdiv{School of Electronic Engineering and Computer Science}, \orgname{Queen Mary University of London}, \orgaddress{\street{Mile End Road}, \postcode{E1 4NS}, \state{London}, \country{United Kingdom}}}
\address[3]{\orgname{Cambridge Digital Humanities}, \orgname{University of Cambridge}, \orgaddress{\street{2 Trumpington Street}, \postcode{CB2 1QA}, \state{Cambridge}, \country{United Kingdom}}}

\corresp[$\ast$]{Corresponding author. \href{mailto:yuxuan.liu@qmul.ac.uk}{yuxuan.liu@qmul.ac.uk}}
\abstract{
Science fiction is widely credited with unsettling gender. This article asks whether that holds at the linguistic level, in how readily gendered reference gives way to its counterpart, a neutral term, or other social category, and whether it separates the genre from general fiction. It does so by masking {\mdseries\itshape man}/{\mdseries\itshape men}, {\mdseries\itshape woman}/{\mdseries\itshape women}, and {\mdseries\itshape person}/{\mdseries\itshape people} across same-period science fiction and general fiction corpora and classifying RoBERTa's five likeliest substitutes by gender. Retention and displacement are permutation-tested over five resampling rounds, and the underlying sentences are read closely. Both corpora cross the gender binary in both directions, but unequally: where {\mdseries\itshape woman} is masked, the model predicts a masculine term about twice as often as it predicts a feminine term where {\mdseries\itshape man} is masked. Each mismatch marks a statistical surprise, a word the corpus did not expect there. Against this shared pattern, science fiction shows greater gender permeability than the general fiction corpus, placing {\mdseries\itshape man} in feminine contexts more often than general fiction while the reverse rate holds steady. Its degendering tendency is stronger too, drawing both gendered terms toward the neutral. Beyond the binary, a masked {\mdseries\itshape woman} also draws substitutes naming a social category and no gender far more often than a masked {\mdseries\itshape man} does. In measuring the gender permeability of {\mdseries\itshape man}, {\mdseries\itshape woman}, and {\mdseries\itshape person} in linguistic distribution and testing whether genre changes it, the study gives science fiction studies linguistic evidence for thematic claims and computational literary studies a pipeline applicable to other category vocabularies, such as race and class.
}


\maketitle

\section{Introduction}

Contemporary discussions of gender increasingly revolve around questions of diversity and inclusion: how people relate to categories such as \textit{man} and \textit{woman}, how those categories make room for nonbinary identifications, and how gender intersects with other axes of identity and group membership (Butler 1990, 1993). This article examines how gender-associated traits are also cued and reinforced in language, where particular lexical choices and recurrent contextual patterns invite familiar, often habitual, interpretations of who is being presented as a \textit{man}, a \textit{woman}, or as more ambiguously gendered (Butler 1990; Eckert and McConnell-Ginet 2003), as well as how these categories admit other forms of social classification. Repeated across many texts, these language practices leave cumulative linguistic traces, informing the expectations that cluster around terms such as \textit{man}, \textit{woman}, and \textit{person}, and shaping the range of contexts in which those terms are likely to be read as fixed, as mutable, or as open to alternative construals (Eckert and McConnell-Ginet 2003).

Masked language modelling makes those traces measurable. The procedure hides one word in a sentence and asks a model trained on large quantities of text which words could stand in its place, returning a ranked set of candidates with a probability attached to each.

Because the model has learned those probabilities from the record of past usage, the candidates it returns for a hidden \textit{man}, \textit{woman}, or \textit{person} report what the record has made most probable in that position. A model of this kind therefore registers both the stabilization of linguistic expectations and the moments at which those expectations are unsettled. Used in this way, it shifts a familiar question in scholarship on gender and science fiction from the level of theme and character to that of language. We ask, when the genre imagines different arrangements of gender, do those rearrangements also reach the language in which they are written?

A growing body of work has documented how language models, and the word embeddings that preceded them, reproduce gender bias, particularly through recurring associations between gendered terms and stereotypical roles or attributes (Bailey \textit{et al.} 2020, 2022, 2025; Caliskan \textit{et al.} 2017; Garg \textit{et al.} 2018). Much of this research, however, has treated gender categories themselves as stable reference points, focusing on how bias accumulates around them and leaving aside whether the categories remain internally coherent in language use. Shifts in the semantic scope of the entities to which bias and stereotypes attach have received comparatively limited scrutiny. That gap opens a set of related questions about categorical fluidity, and about how far patterns of permeability are genre-dependent, with science fiction, compared with general fiction, potentially producing linguistic environments in which the boundaries stabilizing gendered reference are loosened or reconfigured.

Three questions organize the analysis. The first asks how often male, female, and neutral entities (i.e. \textit{person} and \textit{people}) return correct candidate predictions, how often they are displaced by a `misprediction', and what the distribution of these predictions reveals about how gendered reference is organized in literary language. The second asks in which direction the model pushes the displaced terms. We expect masculine predictions for female and neutral words to be far more frequent than the inverse, reflecting the tendency for the masculine to function as the default term for a human being in English usage (Bailey \textit{et al.} 2020, 2022, 2025). We are especially interested in the inversions of that expectation, as well as cases where predictions move beyond gendered reference altogether, towards other forms of social classification and power that are necessarily intersectional in nature (Crenshaw 1989; D'Ignazio and Klein 2020). The third asks whether the distribution found in general fiction persists in science fiction or changes in ways that suggest a distinctive relationship between gendered reference and genre.

\section{Theoretical Framework}

\subsection{Gendered Intelligibility and Human-Reference Terms}

A central concern of feminist and gender theory has been to understand how gender comes to appear as a stable and intelligible feature of social life. Judith Butler's \textit{Gender Trouble} (1990) represents a major intervention in this tradition, challenging the assumption that gender expresses a pre-existing or stable identity (Butler 1990). Rather, Butler argues, gender becomes legible through repeated acts, bodily signs, forms of address, and conventions through which it is enacted. Gender, on this account, has no prior interior essence from which its outward expressions flow; its apparent stability is produced through the repetition of socially regulated forms.

In \textit{Bodies That Matter} (1993), Butler extends this account to discourse itself, arguing that performativity operates as `the reiterative and citational practice by which discourse produces the effects that it names' (Butler 1993). This has implications for the way we understand gendered human-reference terms. To invoke \textit{woman} or \textit{man} in a sentence is to participate, however minutely, in the citational chain through which the meanings of these words accumulate, stabilize, or are strained. Gendered meaning can therefore be understood, at least in part, as sedimented through repeated linguistic usage. Butler's account supplies one of two coordinates for the present analysis: if gendered meanings are sustained through repetition, then the ways in which terms such as \textit{man} and \textit{woman} occur, substitute for one another, or give way to neutral human-reference terms can provide evidence of how those meanings are organized in language.

The second comes from Simone de Beauvoir's \textit{The Second Sex} (1949), in which she famously defined \textit{man} as the absolute human type against which \textit{woman} is figured as the relative and the \textit{Other} (de Beauvoir 2015). It is that specific asymmetry that the analysis also sets out to measure. Locating that asymmetry in a distributional record requires a scale on which referring terms can be placed. This article therefore uses a genericity axis, a spectrum running from fully gender-specific reference at one pole to `generic' human reference at the other. `Generic' in this context refers to language that identifies a human without specifying sex or social gender, as with \textit{person} or \textit{people}. 

\subsection{The Genericity Axis}

Three sets of terms are central to human reference in English: \textit{man}/\textit{men}, \textit{woman}/\textit{women}, and \textit{person}/\textit{people}. The lexicographic record of the three terms (\textit{OED}, s.vv. `man, n.1', `woman, n.', `person, n.'), the history of gendered reference in English (Curzan 2003), and prior distributional research on general English at the scale of hundreds of billions of words (Bailey \textit{et al.} 2022, 2025) establish the starting point for the genericity axis used in this study. This prior research indicates that the three terms do not occupy equivalent positions. \textit{Man}/\textit{men} spans both male-specific and historically generic human reference, giving it a wider range of associated uses; \textit{person/people}, although nominally gender-neutral, occurs within linguistic environments that are themselves often shared by masculine reference; and \textit{woman/women} is more consistently restricted to female-specific reference. The result is an axis whose apparently neutral centre is not insulated from masculine contextual influence.

The axis describes the distributional behaviour of referring terms in the documented record of English usage: it plots words, not persons, and makes no claims about gender identity as lived experience. The axis is also a reduction, since gender does not operate in an isolated dimension and categories such as \textit{man} and \textit{woman} intersect with race, sexuality, class, and other forms of social difference. Intersectional scholarship, perhaps most influentially Kimberlé Crenshaw's work on the mutually constitutive effects of gender and race, has shown how accounts of \textit{woman} can obscure these intersections when they treat gender as a sufficient category in itself (Crenshaw 1989). Catherine D'Ignazio and Lauren Klein extend related concerns into data science, demonstrating how apparently simple classification can reproduce and reinforce wider structures of power (D'Ignazio and Klein 2020). The analysis isolates gendered reference as one dimension in order to examine its linguistic patterning, and where the evidence extends beyond that dimension, as when predictions for \textit{woman} draw in terms associated with race or sexual orientation, the movement beyond the axis is treated as a finding about the language from which the model learned.

Nor is the configuration of the three terms a fixed property of English. It is the accumulated product of the contexts in which these terms have been used. A distributional pattern produced through the accumulation of contextual deployment can, in principle, change when the conditions of deployment change. The distribution documented in general English therefore provides a baseline against which other bodies of language can be compared. The present analysis asks whether literary genre alters that configuration.

\subsection{Genre as a Condition of the Axis}

Science fiction has long been understood as a genre particularly capable of unsettling conventional arrangements of gender. Scholarship on gender in SF has explored this capacity both as a formal feature and as a thematic preoccupation. Helen Merrick identifies SF as a genre whose formal resources are particularly suited to foregrounding texts that disrupt or challenge normative cultural understandings of those roles (Merrick 2003). Brian Attebery's survey of gendered representation in the genre, from \textit{Frankenstein} to late-twentieth-century feminist fiction, similarly shows SF repeatedly incorporating, exploring, and transforming conventional concepts of gender (Attebery 2002). More recent scholarship, including \textit{The Routledge Companion to Gender and Science Fiction}, approaches gender in the genre through intersecting questions of identity, embodiment, race, and species (Yaszek \textit{et al.} 2023).

Computational literary studies has approached the topic of gender in fiction from a different direction. Ted Underwood, David Bamman, and Sabrina Lee measured how easily a character's grammatical gender can be inferred from the words that describe the character, and found the inference growing harder across two centuries (Underwood \textit{et al.} 2018). Their measure registers how sharply description divides characters into two genders. The present study works at the level of the referring term, masking \textit{man}, \textit{woman}, or \textit{person} and classifying the words the surrounding sentence makes probable in its place, which shows how often, and in which direction, each term gives way to another, to a neutral term, or to a term naming another social category.

The distinction matters because existing scholarship establishes that SF can rearrange gender at the levels of theme, character, and plot, but has largely left open whether such rearrangements are also visible in the distributional behaviours of gendered language. In other words, if SF changes the ways in which gender can be imagined or represented, does it also change these exchanges between gendered terms? This article provides a way to test that question. By using masked language modelling to measure which gendered, neutral, and socially marked terms compete for the same linguistic positions, and by comparing those patterns with general fiction, we extend scholarship on gender in SF from representation to linguistic distribution. The contribution is therefore not simply to identify different representations of gender in science fiction, but to measure whether the genre produces a distinct pattern of gendered linguistic expectation.

Such a record has to be read with care. A departure from gendered expectation opens a possibility, but, as Butler's account of gender trouble suggests, what that possibility means depends on how the moment is received, read, and put to use. A woman named where the language leads one to expect a man may unsettle that expectation or merely confirm it. If this example is read as a woman stepping into a man's role, taking on capacities a culture already counts as his, it reaffirms the very order it appeared to breach, leaving the masculine, in the asymmetry de Beauvoir described (de Beauvoir 2015), the measure against which she is read.

What concerns the present study, more narrowly, are the points in literary language at which the settled organization of gendered reference begins to disclose its own conditions. These are moments where a local act of naming presses against a longer history of gendered probability, and where the norms sustaining that history become perceptible as norms – or in Butler's framing, as effects of citation accumulated over time (Butler 1993). The analysis dwells on such moments, holding off from immediate classification so that what is at stake in them can be read; their force emerges through close reading at different scales, beginning with the pressure of the single sentence and extending to the surrounding scene, the work as a whole, the author's recurring patterns, and the wider generic field in which such patterns have accumulated. Read in this way, they become openings for interpretation, sites where gendered reference becomes available for scrutiny before any claim about subversion is made.

\section{Materials and Methods}

\begin{figure*}[p]
\centering
\includegraphics[height=0.80\textheight,keepaspectratio]{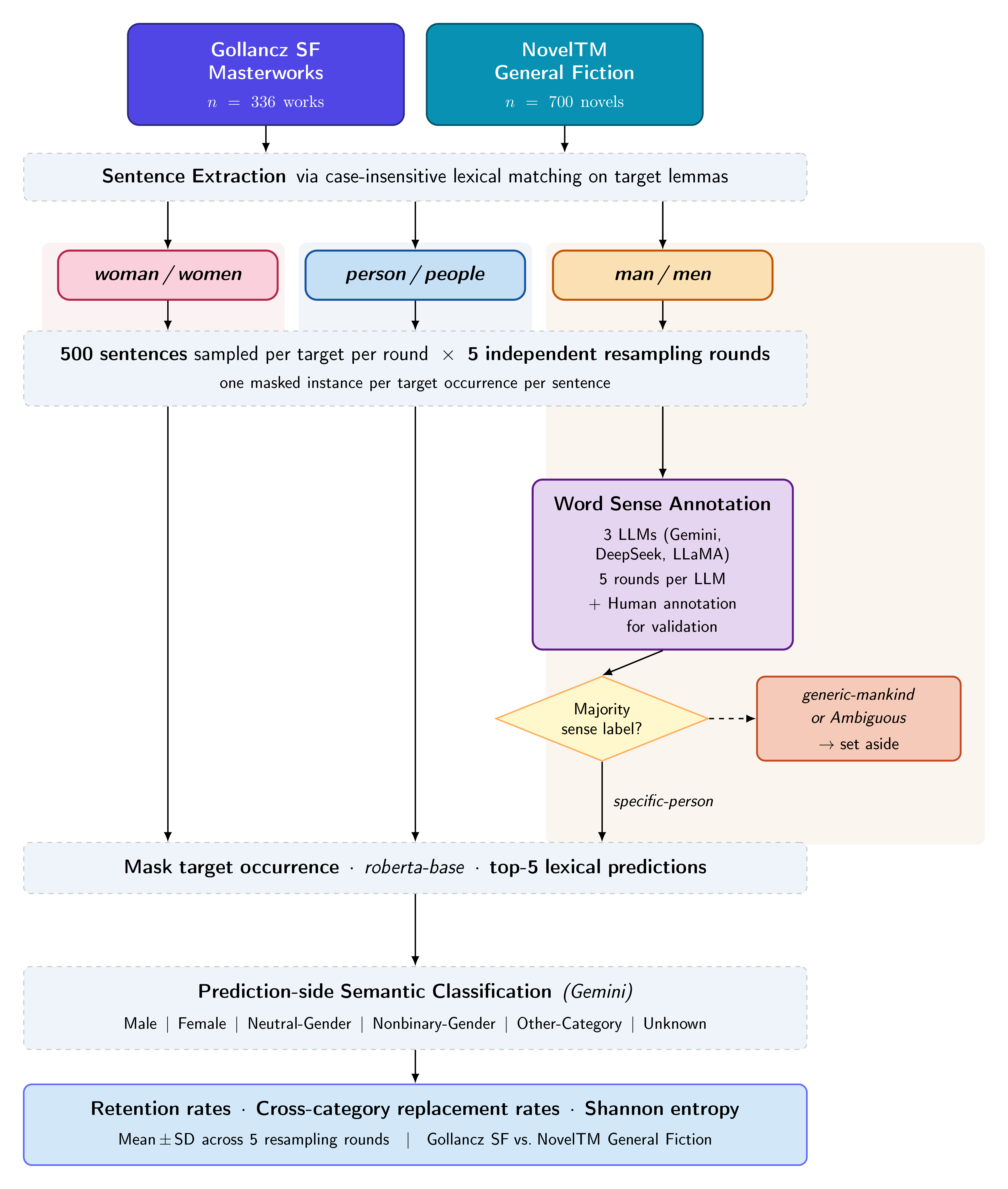}
\caption{Workflow of the gendered human-reference analysis\label{fig1}}
\end{figure*}

The experimental pipeline extends the Mask and Classify workflow introduced by Liu \textit{et al.} (2025), which developed the masked-prediction procedure of the Living with Machines project (Coll Ardanuy \textit{et al.} 2020; Wilson \textit{et al.} 2023), to the analysis of gendered reference. Fig.~\ref{fig1} sets out the full procedure.

The present study draws on two literary corpora. The Gollancz dataset contains 336 works from the Gollancz SF Masterworks series, over 90\% of them dated 1910–2000; the comparison set is drawn from HathiTrust NovelTM (Underwood \textit{et al.} 2020) and comprises 700 randomly sampled Anglophone novels published in the same period. In each corpus, all sentences containing the target lemmas \textit{man}/\textit{men}, \textit{woman}/\textit{women}, and \textit{person}/\textit{people} are collected into lemma-specific pools via case-insensitive matching. From each pool, 500 sentences are randomly sampled per resampling round across five independent rounds. \textit{Man}/\textit{men} receives one further step before sampling. Because the term carries both a generic and a male-specific sense, as set out in Section 2.2, its instances are first annotated for word sense and the generic-mankind uses removed, so that the predictions analysed for this target concern specific-person reference alone. The execution of that step is given in the note.\endnote{The sampled \textit{man}/\textit{men} instances were deduplicated before annotation, ensuring that repeated sentences produced by the resampling procedure were annotated only once, with their labels subsequently mapped back to all corresponding sampled occurrences; instances classified as sense (iii), generic-mankind, were removed before the main sampling step. The annotation was conducted using three large language models in parallel: Gemini 2.5 Pro, DeepSeek-chat, and Llama 3.3 70B. Each model received sentences in shuffled batches of ten, accompanied by a structured prompt specifying the two target senses and requiring JSON output with one entry per token occurrence. Sentences were processed in five independent rounds per model; the majority label across rounds was taken as each model's sense assignment for each occurrence, and the majority label across the three models' assignments was then taken as the final annotation. The full annotation input comprised 4,928 unique \textit{man}/\textit{men} sentences: 2,444 from the Gollancz SF corpus and 2,484 from NovelTM. To assess the reliability of this automated procedure, 100 sentences were selected for independent human annotation by the first author and a computational linguist with expertise in corpus annotation. Human labels were compared against the final majority labels, and the resulting agreement rates were used to estimate the overall accuracy of the automated annotation across the full sentence pool. Whether the three LLMs produced consistent sense distributions was assessed using chi-square tests of independence, with Bonferroni-corrected pairwise comparisons conducted for any model pair where the omnibus test reached significance.}

Within each sampled sentence, target occurrences are masked one at a time, and each masked position is processed with roberta-base (Liu \textit{et al.} 2019) to obtain the top-five lexical predictions with associated probabilities. Each candidate completion is assigned a semantic label via Gemini-based classification under a gender-focused taxonomy mapping predictions to four core categories (Male, Female, Neutral-Gender, Nonbinary-Gender), with two further labels (Other-Category, Unknown) for candidates that name no gender. These final two labels direct us out of the genericity axis into the predictions analysed in Section 5.4, where a masked \textit{woman} draws terms of sexuality, religion, and race.

Retention rates, cross-category substitution rates, and Shannon entropy over the top-five prediction distributions are computed for each target class and reported as means and standard deviations across the five resampling rounds. Retention measures how firmly a term is held in its own category, substitution where it goes when it is not, and entropy how settled the model is about the position; together they give a term's stability, the direction of its exchanges, and the confidence behind them. Entropy is taken over the normalized probability distribution at each masked position; lower entropy indicates a narrow, strongly favoured candidate set while higher entropy signals a broader and more uncertain field of alternatives. Across-corpus differences in retention and replacement are tested using permutation tests with bootstrap confidence intervals; entropy is tested using a two-way ANOVA followed by Bonferroni-corrected Welch post-hoc tests.

\section{Results}

Figure 2 shows a clear ordering in retention rates, which record how strongly each target term is stabilized by its most characteristic category. In NovelTM, the general fiction corpus, \textit{person/people} exhibits the strongest within-category retention, mapping most often to Neutral-Gender (52.4\%), followed by \textit{man} mapping to Male (41.7\%), while \textit{woman} shows the lowest self-retention, mapping to Female at 34.3\%. Gollancz preserves the same ordering, with retention rates of 57.4\%, 39.8\%, and 34.4\% for Neutral-Gender, Male, and Female, respectively.

The ordering is itself asymmetric, \textit{woman} held least securely of the three, and a second asymmetry clarifies how categorical permeability operates inside it, since bidirectional gender crossover is patterned and sharply uneven. For \textit{man}, Male retention is about five times the Female crossover in NovelTM (41.7\% vs. 8.2\%; $\approx$5:1), and remains above four times in Gollancz (39.8\% vs. 9.5\%; $\approx$4:1). For \textit{woman}, Female retention is roughly twice the Male crossover in both corpora (NovelTM: 34.3\% vs. 16.8\%; $\approx$2:1; Gollancz: 34.4\% vs. 17.2\%; $\approx$2:1). The same bidirectional crossover is observed in both targets, yet the imbalance holds across genres: crossover away from Female toward Male is markedly more frequent than crossover away from Male toward Female (Gollancz: $\Delta$ = $-$7.7 pp, 95\% CI [$-$8.6, $-$6.7], p < .001, Cohen's d = 0.43; NovelTM: $\Delta$ $\approx$ $-$8.5 pp, 95\% CI [$-$9.5, $-$7.5], p < .001, Cohen's d = 0.49). This asymmetry reflects biases latent in the data on which RoBERTa was trained: male is often the default subject of human action and the dominant object of narrative attention (Caliskan \textit{et al.} 2017; Eckert and McConnell-Ginet 2003), while female is habitually defined in relation to that male norm, the relative position de Beauvoir set against the masculine Absolute (de Beauvoir 2015). As a result, at the level of distributional prediction, \textit{woman} proves more susceptible to Male-aligned predictions than \textit{man} is to Female-aligned ones.

Entropy adds a complementary view by tracking how concentrated or diffuse the model's expectations are at the masked position (Fig.~\ref{fig3}). Entropy shows that the three target terms are differently ordered across the two corpora in terms of predictability, a pattern supported by a significant category $\times$ corpus interaction (F(2, 15,063) = 8.46, p < .001, $\eta^{2}$ = .0011). In NovelTM, \textit{man} occupies the lowest-entropy position (H = 1.5832) and is significantly lower than both \textit{person/people} (H = 1.6504; t = $-$4.11, p < .001) and \textit{woman} (H = 1.6563; t = $-$4.42, p < .001), indicating that masculine reference forms the most securely anticipated position in the general-fiction corpus. Gollancz reorganizes this pattern. The NovelTM hierarchy centred on \textit{man} loses its dominance, while \textit{person/people} occupies the numerical minimum (H = 1.5931) and shows significantly lower entropy in Gollancz than in NovelTM (H = 1.5931 vs. H = 1.6504; $\Delta$ = $-$.057; t = $-$3.73, p < .001).

\begin{figure*}[!tb]
\centering
\includegraphics[width=0.85\textwidth]{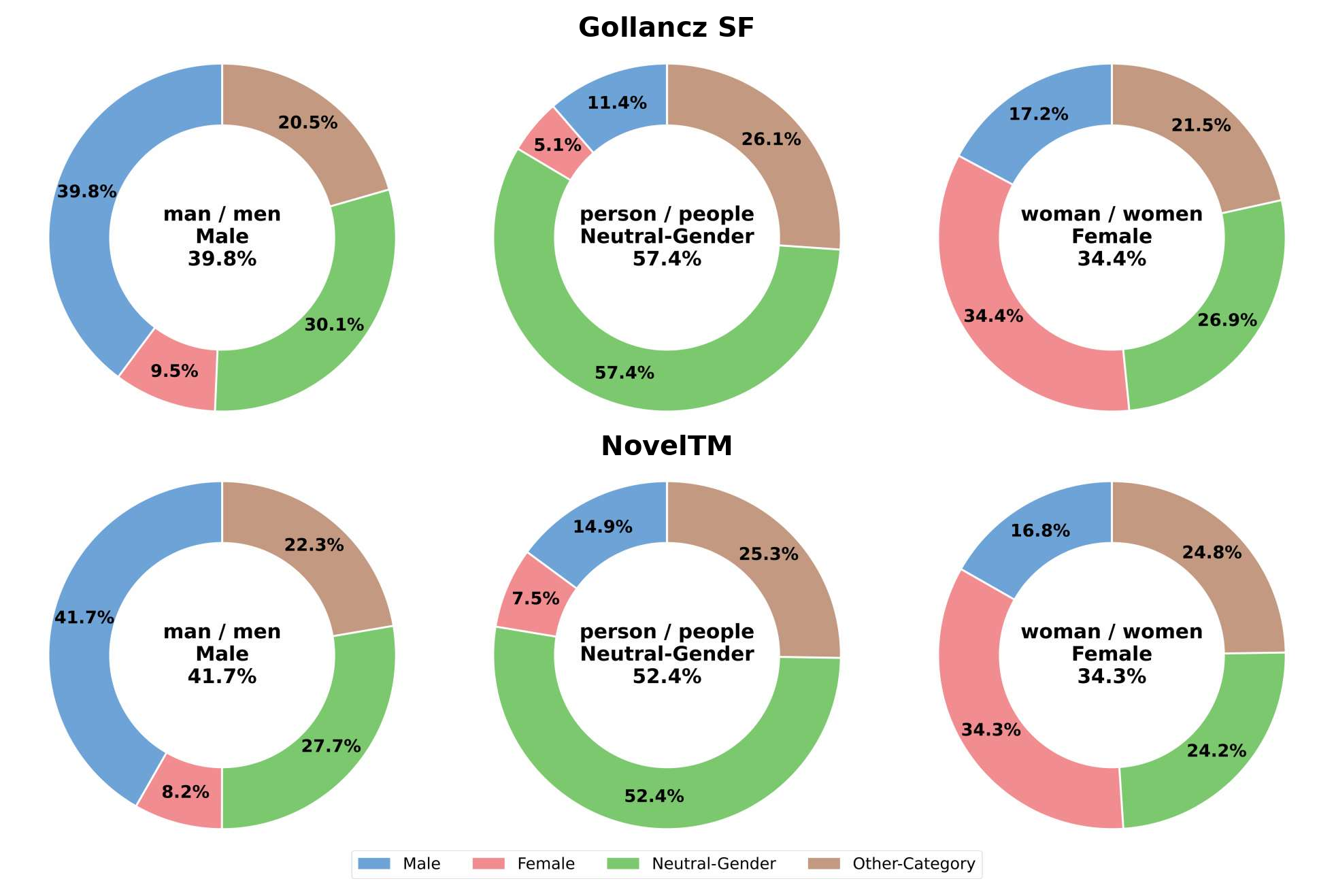}
\caption{Cross-category replacement patterns in the Gollancz SF and NovelTM corpora.\label{fig2}}
\end{figure*}

\begin{figure*}[!tb]
\centering
\includegraphics[width=0.85\textwidth]{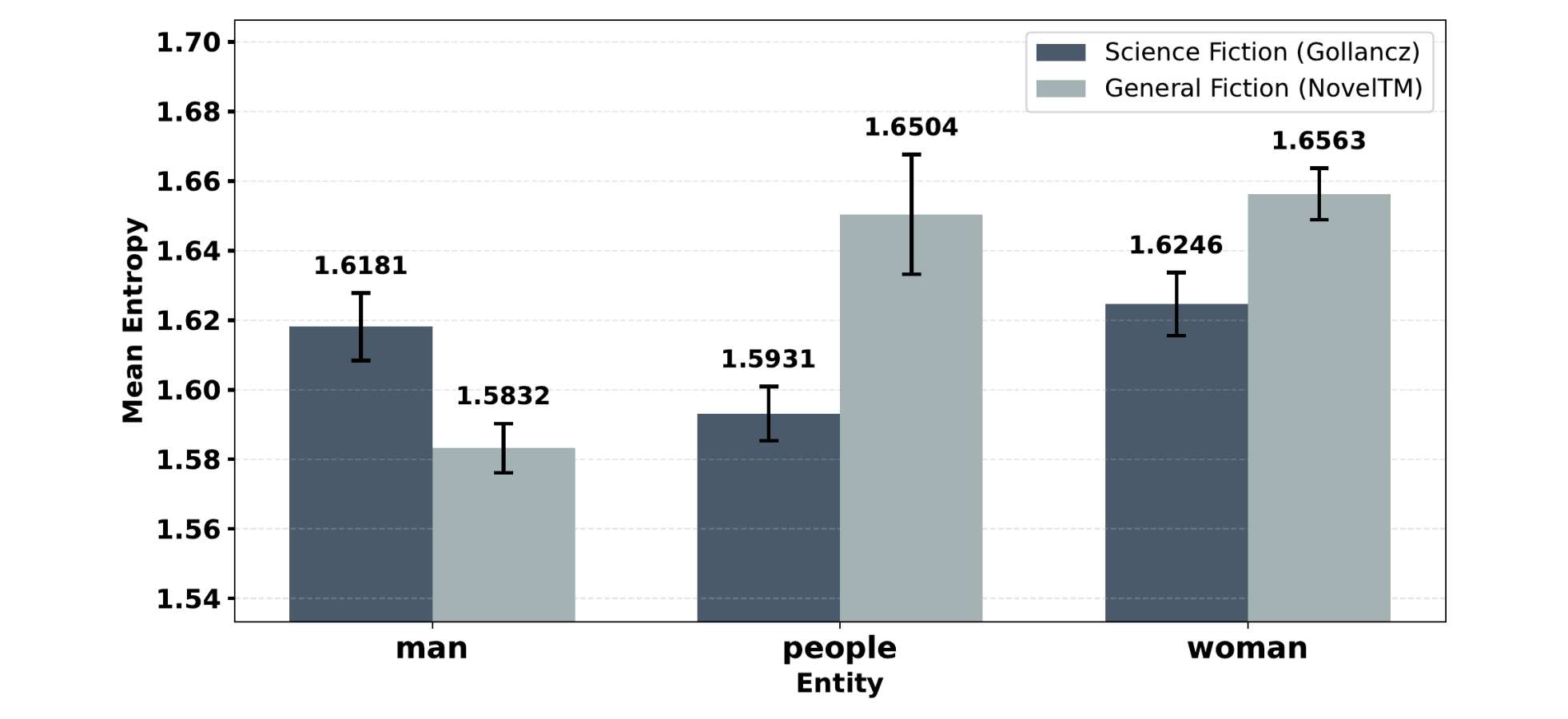}
\caption{Entropy of predictions across target terms in the Gollancz SF and NovelTM corpora.\label{fig3}}
\end{figure*}

This lower entropy coincides with the higher Neutral-Gender retention rate for \textit{person/people} (57.4\% in Gollancz vs. 52.4\% in NovelTM; $\Delta$ = +5.0 pp, p~<~.001), so neutral reference in the science-fiction corpus becomes more stable in both category and prediction: the model is more likely to keep it neutral, and more confident about the lexical possibilities available at the masked position. The entropy changes for \textit{man} and \textit{woman} remain descriptive after correction for multiple comparisons: \textit{man} is numerically higher in Gollancz than in NovelTM (H = 1.6181 vs. H = 1.5832; $\Delta$ = +.035; t = 2.04, raw p = .041), while \textit{woman} is numerically lower (H = 1.6246 vs. H = 1.6563; $\Delta$ = $-$.032; t = $-$2.01, raw p = .045), with neither comparison surviving Bonferroni correction. The replacement results (Fig.~\ref{fig2}) show the complementary side of this reorganization: in Gollancz, the explicitly gendered targets also move more often toward Neutral-Gender than in NovelTM, with higher Neutral-Gender crossover for \textit{man} (30.1\% versus 27.7\%; $\Delta$ = +2.4 pp, 95\% CI [+0.7, +4.2], p~<~.01, Cohen's d = 0.08) and \textit{woman} (26.9\% versus 24.2\%; $\Delta$ = +2.6 pp, 95\% CI [+1.2, +4.1], p~<~.001, Cohen's d = 0.10). The larger pattern is therefore one of neutralization in SF relative to general fiction. 



\section{Discussion}

\subsection{The Masculine Neutral: Generalization, Typification, and Degendering}

Both corpora hold \textit{person} and \textit{people} more securely within the neutral category than they hold either gendered term within its own category (NovelTM: 52.4\%; Gollancz: 57.4\%). When the neutral terms do take a gender, they do so in the masculine direction at roughly twice the feminine rate (NovelTM: 14.9\% vs. 7.5\%; Gollancz: 11.4\% vs. 5.1\%). The strong tendency towards the masculine is not surprising since the proximity of \textit{person} to \textit{man} in aggregate distributional space is already established (Bailey \textit{et al.} 2020, 2022), and, as discussed above, the genericity axis is structurally weighted toward the masculine. What aggregate measures cannot show, however, is the content of the tilt, which type of \textit{man} the nominally neutral term is pulled toward and what in the surrounding linguistic context activates that type. Sentence-level predictions allow us to see this more clearly. The masculine figure deposited in the neutral slot is not uniform, and the attributes that precipitate the substitution differ between the corpora. In Gollancz, sentences containing \textit{person} or \textit{people} that the model reads as masculine tend to involve technical origination, inventive capability, and physical action carried out in groups in spaces of physical instrumentality. In NovelTM, they more often involve honour, professional standing, and the publicly enacted virtues of the gentleman.

These corpus-level tendencies can be seen in individual sentences. In the Gollancz corpus, for example, Pat Cadigan's \textit{Synners} (Cadigan 1991) contains the phrase `the [MASK] who had invented brain sockets', a phrase whose referent the novel identifies as a \textit{woman}. The model predicts \textit{man} with a probability of 0.318, the likelihood the model assigns to that candidate, with no feminine term among its five highest-ranked candidates. In Arkady and Boris Strugatsky's \textit{The Snail on the Slope} (1966; Strugatsky and Strugatsky 1980) the masked \textit{People} `in cardboard masks, holding mine detectors' among ruins is predicted as \textit{Men} with a probability of 0.828. By comparison we might look to the following examples from the NovelTM corpus. `A [MASK] of honor, and not a thief' in \textit{The University Library of Autobiography} (1918) yields \textit{man} with a probability of 0.802, with \textit{knight} and \textit{soldier} in the lower ranked candidates, while the self-description `I am an essentially practical [MASK]' in Harold Bindloss's \textit{His Lady's Pleasure} (Bindloss 1913) returns \textit{man} with a probability of 0.821, with \textit{gentleman} also amongst the candidates. A reader who inferred a man from contexts such as `honor' or `practical' might be challenged for making a gendered assumption. The model makes the same inference because those associations are present in the linguistic record on which its predictions are based. Its predictions, therefore, allow us to observe a pattern of gendered association that might otherwise remain implicit.

These substitutions raise a further question. If \textit{man} can function as a generic human term, as the lexicographic record establishes, then a masked \textit{man} operating as an unmarked human term should generate a prediction of \textit{person} with relative ease: in such contexts the two terms perform similar distributional work. The corpora do indeed show this substitution at a measurable rate (NovelTM: 27.7\%; Gollancz: 30.1\%). More strikingly, however, \textit{woman} generates a neutral prediction at a rate only a few points lower (NovelTM: 24.2\%; Gollancz: 26.9\%). Given the historical asymmetry between masculine and feminine reference, we might expect \textit{woman} to be substantially further from the generic position. The rates alone would suggest that the two terms enjoy a similar freedom. Looking at the sentences themselves, however, reveals that the two terms reach neutral reference through different linguistic routes.

Tables 1 and 2 gather illustrative examples of those routes. The examples comprise six instances in each direction, selected from the crossover sets of the two corpora. They are intended to illustrate the linguistic contexts in which the substitutions occur. Within these examples, \textit{man} and \textit{woman} reach neutral reference in different grammatical and semantic contexts. Where \textit{man} yields \textit{person} (Table~\ref{tab1}), the surrounding construction typically involves a frame that counts, orders, or ranks its referent through position in a series. The substitution therefore produces a genus: \textit{person} names the whole class from which the individual is drawn, `the first person who', `the last person in the world', `one person can do'. Where \textit{woman} yields \textit{person} (Table~\ref{tab2}), by contrast, the substitution more often produces a specimen or typification: \textit{person} names the object being sorted into a type, `the type of person who', `the sort of person who', `one of those people who'. Therefore, the two gendered terms differ in the route by which they reach the neutral as well as in their distance from it, the masculine arriving through generalization and the feminine through typification, so the asymmetry survives inside the very movements that appear to suspend it, and even where gender falls away from reference, it falls away differently by gender.

The two routes hold in both corpora; what changes is their frequency. Science fiction produces neutral predictions for both \textit{man/men} and \textit{woman/women} more often than general fiction: 30.1\% versus 27.7\% for \textit{man/men}, and 26.9\% versus 24.2\% for \textit{woman/women}. These differences are modest in absolute terms, but both are statistically significant, and both favour the SF corpus. In other words, the SF corpus more often places gendered terms in contexts where a neutral human-reference term could reconstruct the sentence. The terms remain explicitly gendered, but their immediate contexts make gender less necessary for identifying the referent; what matters more is that the referent is legible as a person, a human individual, or a member of some broader class.

The entropy results reported in Section 4 point in the same direction. The terms associated with the most confident predictions shift from \textit{man} in NovelTM to \textit{person} and \textit{people} in Gollancz. The substitution and entropy measures together suggest a broader pattern of genre-level degendering: science fiction more often places gendered terms in contexts where a neutral term could stand, while also making neutral terms more predictable. Crucially, however, this increased neutralization does not remove the asymmetry between masculine and feminine reference. Instead, it changes how often gender can be set aside while leaving the different routes by which \textit{man} and \textit{woman} reach neutral reference intact.

\begin{table*}[!tb]
\caption{\textit{Man}/\textit{men}-to-\textit{person}/\textit{people} predictions\label{tab1}}
\centering\footnotesize\begin{tabular}{@{}p{0.075\textwidth}p{0.185\textwidth}p{0.50\textwidth}p{0.10\textwidth}@{}}
\toprule
\textbf{Corpus} & \textbf{Work} & \textbf{Sentence ([MASK] = \textit{man/men})} & \textbf{Rank-1 prediction (p)} \\
\midrule
Gollancz & H. G. Wells, \textit{The First Men in the Moon} (Wells 1901) & `the last [MASK] in the world' & \textit{person} (0.727) \\
Gollancz & James Morrow, \textit{This Is the Way the World Ends} (Morrow 1986) & `there's a limit to what one [MASK] can do' & \textit{person} (0.693) \\
Gollancz & Gene Wolfe, \textit{Shadow and Claw} (Wolfe 1994) & `I was a whole [MASK] again' & \textit{person} (0.799) \\
NovelTM & Hugh Walpole, \textit{The Duchess of Wrexe} (Walpole 1914) & `the first [MASK] who breaks her trust' & \textit{person} (0.819) \\
NovelTM & Arthur Christopher Benson, \textit{Father Payne} (Benson 1916) & `If I see another [MASK] behaving well where I should behave ill' & \textit{person} (0.785) \\
NovelTM & Mary Fisher, \textit{The Treloars} (Fisher 1917) & `the shavings of other [MASK]'s minds' & \textit{people} (0.989) \\
\botrule
\end{tabular}
\end{table*}\begin{table*}[!tb]
\caption{\textit{Woman}/\textit{women}-to-\textit{person}/\textit{people} predictions\label{tab2}}
\centering\footnotesize\begin{tabular}{@{}p{0.075\textwidth}p{0.185\textwidth}p{0.50\textwidth}p{0.10\textwidth}@{}}
\toprule
\textbf{Corpus} & \textbf{Work} & \textbf{Sentence ([MASK] = \textit{woman/women})} & \textbf{Rank-1 prediction (p)} \\
\midrule
Gollancz & Joseph O'Neill, \textit{Land Under England} (O'Neill 1935) & `the type of [MASK] who would normally have an influence over him' & \textit{person} (0.868) \\
Gollancz & Philip K. Dick, \textit{The Three Stigmata of Palmer Eldritch} (Dick 1965) & `the sort of [MASK] who has active imagination' & \textit{person} (0.699) \\
Gollancz & Karen Joy Fowler, \textit{Sarah Canary} (Fowler 1991) & `[MASK] of a certain age are so prone to this' & \textit{People} (0.614) \\
NovelTM & Louis Joseph Vance, \textit{The Fortune Hunter} (Vance 1910) & `one of those [MASK] who can rise to any emergency' & \textit{people} (0.621) \\
NovelTM & Martha Dickinson Bianchi, \textit{The Sin of Angels} (Bianchi 1912) & `another sort of [MASK] who take it up because they are idle and miserable' & \textit{people} (0.670) \\
NovelTM & George Gibbs, \textit{Madcap} (Gibbs 1913) & `the kind of [MASK] who seeks to pry into the affairs of other people' & \textit{person} (0.834) \\
\botrule
\end{tabular}
\end{table*}

\subsection{Woman as Man: Capability on Masculine Ground}

Feminine-to-masculine crossings share some observable trends. Table~\ref{tab3} gathers six illustrative examples, three from each corpus. Masked \textit{woman}/\textit{women} receives a masculine first-rank prediction in 16.8\% of NovelTM positions and 17.2\% of Gollancz positions. This is in comparison to 34.3\% (NovelTM) and 34.4\% (Gollancz) of cases where it is retained as feminine, and against 8.2\% (NovelTM) and 9.5\% (Gollancz) for the reverse crossing, where \textit{man}/\textit{men} receives a feminine prediction. The six examples place women in contexts of expertise, professional activity, public authority, and practical capability: teaching biology, practising medicine in a domestic comedy, and addressing political meetings in the suffrage decade. In each case, the model reverts to the masculine term most strongly associated with such contexts in the literary record.

One of the six, however, differs from the others. Whereas the other five describe a capable woman inside a social world whose gender arrangements remain conventional, the sentence from Ursula Le Guin's \textit{The Word for World Is Forest} (Le Guin 1972) describes a society in which civic governance has been assigned to women. The narrative follows the logging of Athshe, a forested planet colonized for timber by settlers from an Earth that has cut down its own, with the sentence reporting how the Athsheans ran their towns before those settlers arrived. With \textit{women} masked, masculine predictions absorb nearly the whole probability mass, with \textit{men} (p = 0.109) and \textit{gentlemen} (p = 0.072) appearing among the leading candidates (the highest proportion in the table).

Darko Suvin describes such an operation as estrangement: that is, as presenting the apparently given as contingent and potentially otherwise (Suvin 1979). Le Guin applies this operation to gender, imagining women as the civic governors of the society she describes. Yet the model's prediction is pulled back towards the gendered associations established in the wider linguistic record: faced with the sentence `ran the cities and towns' the model predicts not \textit{women} but the figures that history has conventionally placed in charge. The presence of \textit{gentlemen} in the candidate set sharpens the picture, since the term names civic authority in its social dress, maleness joined to rank and public standing. Le Guin's proposition names a world organized otherwise; the model names the world the training corpus was assembled from. The gap between the two is what the method is built to find. 

Sentences of this kind are rare in the Gollancz crossover set, despite science fiction's documented readiness to rearrange the sexes (Merrick and Tuttle 2022). The crossings mostly attach masculine-coded capability to an individual woman while leaving the wider social arrangements of the fictional world intact at the level of the language: whatever the unfamiliarity of the worlds science fiction imagines, the words in which its crossings are written stay close to those of general fiction.

Within science fiction, this pattern has a history of its own. Reversed sexual arrangements and women occupying conventionally masculine roles are among the genre's established estranging devices (Merrick and Tuttle 2022). A device so well established raises the question of whether repetition wears it out. Joanna Russ argued in 1971 that a genre's characteristic plots and devices can lose their force through repeated use: what begins as an unfamiliar possibility can become plausible, conventional, and eventually little more than an ornamental or self-referential feature of the genre (Russ 1971). For a genre concerned with producing novelty, the repetition that establishes a device can therefore also diminish its capacity to surprise.

This creates an interesting tension with the linguistic patterns being examined in this article. Novelty is spent within a reading community, each reuse of the reversal depleting the surprise on which its estranging force depends, whereas resignification, on the Butlerian account set out above, is earned across the whole citational field. Our analysis would suggest that science fiction has not repeated gender reversals at a frequency sufficient to shift the broader linguistic expectations surrounding \textit{woman}. Feminine-to-masculine crossing occurs at almost the same rate in the two corpora (17.2\% in Gollancz; 16.8\% in NovelTM), and, where science fiction reverses the gender arrangements of its fictional worlds, the model still strongly prefers the masculine term. Whatever the reversal has spent inside science fiction, it has not spent enough to move what the model expects: women running cities remains improbable in the aggregate of English prose.

Here it is worth pointing out that these results are measured across the corpora as a whole rather than by decade, because dividing the relatively small samples into individual periods would make the estimates too unstable. With larger datasets, however, the same procedure could be applied diachronically to ask whether the predictive strength of particular gender reversals changes as they become more or less conventional. Such a temporal approach would make it possible to track whether the surprise associated with a particular device declines as its linguistic contexts become more familiar, which would intersect with emerging work on historical and other forms of contextualized language models (Beelen and van Strien 2022; Underwood \textit{et al.} 2025; Fittschen \textit{et al.} 2026), and with humanities-led calls for such models (Klein \textit{et al.} 2026).

This tension also bears on what science fiction seeks to achieve through estrangement (Suvin 1979). The linguistic evidence presented in this article suggests a paradox: repetition on the scale that would re-weight the contexts of \textit{woman} is the repetition that spends the device. In other words, a reversal written often enough to shift what English expects would by then have forfeited the value the genre assigns it, while a reversal kept rare enough to estrange stays too sparse to shift anything. 

But even if such patterns were to become more common, that would not necessarily alter the linguistic position of \textit{woman}. A departure from \textit{woman} moves a woman onto masculine ground from which she departs largely unchanged. A revaluation of the feminine position would instead be visible in what the contexts of \textit{woman} themselves attract, in the term's own neighbourhood. If masculine capability becomes increasingly available as a prediction for \textit{woman}, that tells us that women are being associated with capabilities historically coded as masculine; it does not, by itself, tell us that the meanings and associations surrounding \textit{woman} have been transformed. Russ identified a related asymmetry in her account of the suppression of women's writing: women's entry into masculine-coded domains may be celebrated while the feminine domain remains undervalued (Russ 1983). R. W. Connell records a related objection to degendering, arguing that it asks women to become like men, making sameness with the masculine the condition of equal standing (Connell 1995). The crossover rates that we report give that objection a distributional form. In these corpora masculine-coded capability reaches \textit{women} relatively readily, but it does so by moving the feminine term onto linguistic ground already marked masculine, in a pattern consistent with the broader structures of masculine domination described by Pierre Bourdieu (Bourdieu 2001). A different pattern would require the linguistic ground itself to change: capability and authority becoming associated with contexts of woman that are no longer already indexed as masculine, and with forms of reference that extend beyond the gender binary. In that case, the words surrounding \textit{woman} would change alongside the roles it assigns her. Inventing a different social arrangement and inventing the words for it are separate undertakings, and the near-identical crossover rates between the science fiction and general fiction corpora suggest that science fiction has gone further with the first than with the second. Crossings that land on feminine ground run in the other direction, and it is there that the two corpora part.

\begin{table*}[!tb]
\caption{\textit{Woman}/\textit{women}-to-\textit{man}/\textit{men} predictions\label{tab3}}
\centering\footnotesize\begin{tabular}{@{}p{0.075\textwidth}p{0.175\textwidth}p{0.51\textwidth}p{0.10\textwidth}@{}}
\toprule
\textbf{Corpus} & \textbf{Work} & \textbf{Sentence ([MASK] = \textit{woman/women})} & \textbf{Rank-1 prediction (p)} \\
\midrule
Gollancz & Olaf Stapledon, \textit{Sirius} (Stapledon 1944) & `This quick-witted and sensitive young [MASK] taught biology, but was also a lover of literature' & \textit{man} (0.734) \\
Gollancz & Ursula K. Le Guin, \textit{The Word for World Is Forest} (Le Guin 1972) & `In all the Forty Lands, [MASK] ran the cities and towns, and almost every town had a Men's Lodge' & \textit{men} (0.747) \\
Gollancz & Christopher Priest, \textit{The Affirmation} (Priest 1981) & `Two other young [MASK], both wearing the smart company outfit, were working behind desks: one was speaking on the telephone, the other was typing' & \textit{men} (0.685) \\
NovelTM & Edward S. Moffat, \textit{Hearts Steadfast} (Moffat 1915) & `You are a [MASK] whose mind deals in big things first, and you'd go through hell fire to do them' & \textit{man} (0.647) \\
NovelTM & Jerome K. Jerome, \textit{All Roads Lead to Calvary} (Jerome 1919) & `Clever, brilliant young [MASK], accustomed to sweep aside all opposition with a blaze of rhetoric' & \textit{men} (0.610) \\
NovelTM & Meredith Nicholson, \textit{Blacksheep! Blacksheep!} (Nicholson 1920) & `Dr. Reynolds is a skilful young [MASK], and of course Isabel and Ruth are strong, vigorous girls' & \textit{man} (0.702) \\
\botrule
\end{tabular}
\end{table*}

\subsection{Man as Woman: Feminization within and beyond Gendered Conventions}

Feminine-aligned candidates appear behind a masked \textit{man/men} in 8.2\% of NovelTM (general fiction) cases and 9.5\% of Gollancz science fiction cases. This is the rarer direction of crossing. Male retention for \textit{man} falls from 41.7\% in NovelTM to 39.8\% in Gollancz ($\Delta$ = $-$1.9 pp, p < .05), and the gap between retention and crossover narrows from 33.5 percentage points to 30.3. The genre difference, therefore, is concentrated in the direction the wider record most resists. Not every feminine prediction for a masculine term, however, marks the same kind of crossing. The aggregate figure conflates cases in which a man is placed within a situation the literary record has statistically associated with female subjects, and cases in which the conditions that ordinarily produce masculine reference have been modified, suspended, or dissolved. Distinguishing between these possibilities requires us to recover the social and linguistic conditions presented by each sentence.

In examples of the former case, the gendered arrangement of the fictional world remains conventional while the male figure alone moves into a situation or register associated with female subjects. If we turn first to the general fiction corpus, in Mary Jane Holmes's \textit{Family Pride} (Holmes 1867) `the strong [MASK], no longer able to bear up, wept like a very child', returns \textit{woman} (p = 0.695) for the masked \textit{man}. In C. N. and A. M. Williamson's \textit{Everyman's Land} (Williamson and Williamson 1918), `a [MASK]'s voice caught it up, a voice of velvet, a voice of the heart', returns \textit{woman} (p = 0.504) ahead of \textit{girl} (p = 0.045). We see similar examples in the Gollancz corpus. In Lucius Shepard's \textit{Life During Wartime} (Shepard 1987), `he thought the old [MASK]'s frail whisper must be commenting upon him' returns \textit{woman} at 0.553; and in Philip K. Dick's \textit{Time Out of Joint} (Dick 1959) `The [MASK]'s voice rose in hysteria' yields \textit{woman} (p = 0.214) ahead of \textit{man} itself (p = 0.146), with \textit{girl} (p = 0.095) following. The predictions reflect a cluster of associations that English prose has historically attached to female subjects: frailty, bodily deterioration, and emotional dependence.

The foregoing examples can be understood as feminizing the masculine figure while leaving the wider gender order intact. By contrast, the second category of examples might be characterized as seeking to shift the poles themselves. The following examples redistribute who occupies the position of aesthetic object, who is subject to another's possessive regard, and which side of these exchanges is marked as feminine. They can do so because masculinity is not carried by the referent alone; it depends on the conditions a social world provides. When those conditions are modified or withdrawn, a male figure may no longer sustain the masculine register the model expects. Bourdieu describes the habitus as a durable embodied system of dispositions generated through sustained exposure to social order (Bourdieu 2001). In this account, masculinity is maintained through institutions, practices, bodily styles, and forms of recognition; Michael Kimmel similarly describes the validation of American manhood as a process that takes place through the approving regard of other men (Kimmel 1996). Under altered conditions, masculine and feminine qualities can become less securely attached to male and female persons, and can instead be distributed across bodies, structures, practices, and discourses (Butler 1990, 1993).

The Gollancz crossover set contains several examples in which science fiction modifies conditions at the level of the fictional world. Each \textit{man}-to-\textit{woman} prediction of this kind is the training data's habitus reimposing its accustomed conditions on a sentence that has suspended them, the record supplying the completion its own conditions would have required. Such predictions therefore are not simply a response to the referent's sex, but more accurately a response to the social and aesthetic conditions in which the referent is presented. Ian McDonald's \textit{Desolation Road} (McDonald 1988), a chronicle of a Martian settlement in which embodiment can be revised by machine intelligences figured as angels, supplies one example. In the sentence `Its face and torso were those of an extraordinarily beautiful young [MASK]', the masked \textit{man} yields \textit{girl} (p = 0.104), \textit{creature} (p = 0.073), \textit{man} (p = 0.054), and \textit{child} (p = 0.036) behind the feminine rank one. The description surveys the body component by component, foregrounding its aesthetic appeal. The model consequently reassigns the figure to the female category that such an inventory has historically presupposed. The next sentence in McDonald's novel makes the revision more explicit: `\,``Or neither'', said the angel, and repeated the facial thaw trick, melting into an extraordinarily beautiful young [MASK] of indeterminate gender'. Here the masked \textit{person} yields \textit{woman} (p = 0.501), \textit{man} (p = 0.172), \textit{girl} (p = 0.095), \textit{creature} (p = 0.054), and \textit{lady} (p = 0.041). Female-aligned candidates therefore absorb roughly two thirds of the probability mass against the text's explicit assertion of indeterminacy. The model's response shows how strongly beauty remains associated with feminine reference in the linguistic material on which it was trained. McDonald's world permits gender to be fluid, elective, and dissolvable, but the predictions are unable to break the associations of the Butlerian citational chain.

Joanna Russ's \textit{The Female Man} (Russ 1975) provides a related example by placing male bodies in the position of desired object. The sentence comes from the novel's female narrator questioning the terms of her own desire: `Does it count if you love [MASK]'s bodies but hate [MASK]'s minds?' The two occurrences of \textit{men} are masked in turn. At the first position, referring to the bodies that are loved, the leading candidate is \textit{women} (p = 0.521), with \textit{men} close behind (p = 0.471). At the second, referring to the minds held in contempt, \textit{men} is retained securely (p = 0.769), with \textit{women} well behind (p = 0.221). The masking therefore separates two associations that the sentence deliberately brings together: the body available to desire is drawn towards the feminine, while the mind subject to judgement remains masculine. Russ's narrator questions her own desire; the candidate sets show which parts of that configuration the linguistic record is prepared to grant to a man and which it withholds.

These examples show that feminization does not produce a single, stable alternative to masculine reference. A man may be feminized by being placed in a context associated with women, or by occupying a position whose gendered terms have been redistributed. In both cases, the movement remains legible along the gender axis. But the predictions reveal, for \textit{woman} especially, a further possibility: the term can also attract candidates that do not simply exchange one gender for another, but invoke other forms of social classification. The next section follows this lateral movement, asking what becomes visible when the gender axis opens into other dimensions of sexuality, race, religion, and social belonging.

\subsection{Woman and the Wider Field of Social Difference}

The candidate sets for \textit{woman/women} contain another striking pattern. Alongside gendered and neutral human-reference terms, they sometimes include words marking sexuality, race, religion, and other forms of social classification. Specifically, the corpora returned clusters of candidates associated with sexuality or sexual orientation, such as \textit{homosexuals}, \textit{gays}, and \textit{lesbian}, as well as demographic and identity categories such as \textit{Jews}, \textit{blacks}, \textit{Muslims}, and \textit{immigrants}. These predictions do not simply name another gender. They draw on vocabularies shaped by histories of stigma, legal prohibition, administration, and contested social belonging.

The pattern is rare in absolute terms and differs somewhat between the two corpora. Across the five resampling rounds, sexuality- or sexual-orientation-based mispredictions (\textit{homosexual/s}, \textit{homosexuality}, \textit{gay/s}, \textit{lesbian/s}) enter the candidate sets of 13 masked \textit{woman/women} positions in Gollancz, out of 2,418 positions in all, and of 8 in NovelTM out of 2,554. Because a single candidate set can contain more than one such term, these correspond to 15 predictions in Gollancz and 9 in NovelTM. The wider vocabulary of marked social categories (\textit{Jew/s, Muslim/s, black/s, immigrant/s, Christian/s}) appears at 27 Gollancz positions and 19 NovelTM positions. These terms occur at every level of the candidate set, from first to fifth rank. The counts are higher in Gollancz, but the frequencies are too low to carry a genre-level claim. Their importance rests instead on the shared pattern: the candidate sets for women include forms of social classification that are largely absent from the corresponding sets for men.

In NovelTM, no sexual-minority term enters any candidate set for \textit{man/men}; in Gollancz there are three appearances, each arising where the sentence itself stages a same-sex pairing (`Then remember you're both [MASK]' and `That we're both [MASK]', drawing \textit{gay} in the lower ranks; `I never knew about [MASK] and women doing things like that', drawing \textit{lesbians}). These three cases are too few to support a positive claim about the contexts in which such vocabulary arises. They do, however, support a more limited observation: across 2,250 masked \textit{man/men} positions in Gollancz and 2,309 in NovelTM, sexual-minority vocabulary does not enter through the same wider discourse of classification, social standing, and exclusion that gathers around \textit{woman}. The fact that \textit{woman} keeps this company while \textit{man} largely does not is itself a historical formation.

Michel Foucault's account of classification helps illuminate this asymmetry. Modern forms of knowledge, he argues, do not simply describe pre-existing kinds of people but produce categories through which populations become objects of knowledge and regulation (Foucault 1978). We can extend this insight to the linguistic pattern observed here: \textit{man} can function as an implicit human standard against which more explicitly classified groups are distinguished, and therefore appear less marked precisely because its own classification remains unspoken.

The corpus holds the sediment of this asymmetry. It carries the categories that these classificatory practices produced, together with the discursive associations that assembled around them, and the way those categories became entangled. This is where Crenshaw's account of intersectionality becomes useful: race, gender, sexuality, religion, and other forms of classification acquire meaning partly through their relations to one another (Crenshaw 1989). The predictive neighbourhood of \textit{woman} is one place where those relations become visible.

George Eliot's \textit{The Mill on the Floss} (Eliot 1860), in the NovelTM general fiction corpus, provides an example: `And the men at St. Ogg's were not all brave, by any means: some of them were even fond of scandal, and to an extent that might have given their conversation an effeminate character, if it had not been distinguished by masculine jokes, and by an occasional shrug of the shoulders at the mutual hatred of [MASK]'. The masked word, \textit{women}, returns first \textit{women} (p = 0.624) and \textit{men} second (p = 0.037), before \textit{homosexuals} appears at the third rank (p = 0.026) and \textit{homosexuality} at the fourth (p = 0.018). The lower-ranked candidates reveal additional associations, visible only at the level of the distributional neighbourhood. Reading the five candidates together makes it possible to see \textit{homosexuals} and \textit{homosexuality} as part of the discourse surrounding the masked position, even though neither displaces \textit{women} as the most likely completion. The candidate set thus provides evidence not only about the word the model expects, but also about the other categories that occupy the same local linguistic space.

The word `mutual', in Eliot's sentence, is key to understanding this linguistic space. The men share a hatred, and that shared hatred constitutes them as a group whose solidarity is performed through collective contempt. The rejection of women operates here as a homosocial credential: it helps establish the standing that each man claims in the others' regard. In the medical and juridical writing of the period, effeminacy in a man was also becoming an important index through which the emergent category of the male homosexual was identified. The two forms of disavowal – of women and of erotic relations between men – could therefore operate together. Eve Kosofsky Sedgwick described this structure in \textit{Between Men}, where the homosocial bond is secured by this paired refusal, whose meanings reinforce one another (Sedgwick 1985). The candidate set recovers \textit{homosexuals} and \textit{homosexuality} from a sentence about the contempt of men for women because both terms have occupied related discourse positions within that history of classification and exclusion.

Similar examples appear in the Gollancz corpus. In H. G. Wells's \textit{The Shape of Things to Come} (Wells 1933), the sentence `Outside that sexual vindication, [MASK] at that time had little to contribute to the solution of the world's problems, and as a matter of fact they contributed nothing' returns \textit{homosexuals} at the first rank (p = 0.259), \textit{women}, the word Wells wrote, second (p = 0.206), and \textit{Jews} third (p = 0.143). Once again, the prediction places \textit{women} within a wider vocabulary of groups whose social position is defined through processes of exclusion and differentiation. Rather than simply identifying women with these other categories, the candidate set suggests that they occupy related positions within a discourse that sorts some groups as socially peripheral to the dominant order.

In Gwyneth Jones's \textit{Castles Made of Sand} (Jones 2002), the narrator observes of a group she is working with: `They're not homophobic, they even believe in civil rights for [MASK], and they don't give a toss for my dissolute lifestyle, if I serve their purpose'. Masking \textit{women} here returns \textit{gays} at the second rank (p = 0.201) and \textit{homosexuals} at the third (p = 0.108), behind \textit{women} (p = 0.359). Where Wells's exclusionary prose pushes the model towards homosexuals, Jones's advocacy-oriented prose pushes the model towards \textit{gays}, a term with its own complex history of usage, reclamation, and political identification (Chauncey 1994; Weeks 1977). The two passages therefore differ in tone and political purpose, but their candidate sets reveal a shared categorical grammar: the writing that assigns \textit{woman} a place also brings into view the standing of sexual and racial minorities. The surrounding politics may change, but the categories remain connected within the linguistic patterns inherited by the model.

These examples also point to a broader methodological possibility. The value of the candidate set is not exhausted by identifying the model's most likely prediction. The full distribution provides a way of tracing the linguistic field around a word: the terms that almost displace it, the categories that repeatedly appear alongside it, and the associations that may remain implicit in the text itself. In this sense, the model can be used not only to locate moments of textual friction, where a prediction sharply diverges from the word on the page, but also to work backwards from those predictions to the broader currents of discourse in which a term has acquired its meanings. As such, we see here a way to connect computational approaches back into the practice of literary criticism. A close reading can identify the explicit argument of Wells's or Jones's passage; the candidate set adds another layer, showing how the language of those passages sits within a wider network of historically sedimented associations. What emerges is not a hidden meaning that the model has somehow discovered, but a distributional structure that becomes perceptible when the model's alternatives are treated as evidence. The method therefore allows us to move between the local and the systemic: from an unexpected prediction at a particular sentence to the wider linguistic relations that make that prediction possible.

\section{Conclusion}

This study has used masked language modelling to examine gender not simply as a property represented in literary texts, but as a set of linguistic expectations against which those texts can depart. Three findings emerge. First, the genericity axis documented for general English is visible in both literary corpora. \textit{Person}/\textit{people} tends towards the masculine when it takes a gender, while \textit{man} enters neutral human reference through generalization and \textit{woman} through typification. The binary is crossed unevenly: roughly one in six masked occurrences of \textit{woman} admits a masculine reconstruction, compared with fewer than one in ten masked occurrences of \textit{man} admitting a feminine one. The two routes into neutral reference differ in kind. Gender may disappear from the surface of a sentence without erasing the asymmetry between the terms from which that neutrality is reached.

Second, science fiction modifies the axis at specific points while leaving its overall linguistic weighting largely unchanged. Neutral human reference becomes more predictable, and the strongest predictive concentration shifts away from \textit{man} towards \textit{person/people}. The genre also permits more movement from masculine to neutral reference and a somewhat greater permeability of the binary from the masculine side, while the movement from \textit{woman} to \textit{man} remains remarkably stable across the two corpora. Science fiction can therefore unsettle the organization of gendered reference without necessarily redistributing its underlying asymmetries. Its imaginative reordering of gendered worlds and its reorganization of the linguistic associations through which those worlds become predictable are related, but distinct, processes.

Third, the prediction neighbourhood of \textit{woman} extends beyond the gender binary into other forms of social classification. Sexuality, race, religion, and other marked categories appear among its candidate sets in ways that are largely absent from those surrounding \textit{man}. These associations should not be read as evidence that the categories are equivalent, nor as direct measures of authorial intention. Rather, they show how gendered reference participates in a wider linguistic history in which social categories acquire meaning through their relations to one another. The candidate set makes this relational structure perceptible: not as a hidden meaning behind the text, but as a distribution of linguistic possibilities surrounding a word.

Taken together, these findings suggest a way of bringing computational and literary modes of reading into a more productive relation. The model's distributions do not explain a text in place of interpretation; they provide another scale at which literary language can be read. A sentence can be considered against the wider linguistic patterns that make it predictable or surprising, while aggregate patterns can in turn be returned to individual passages and historical contexts. What matters is not choosing between the singularity of close reading and the scale of computational analysis, but using each to make questions possible at the other scale.

As such we have shown how the Mask and Classify procedure (Liu \textit{et al.} 2025) is more than just a technique for classifying gender; it is a general framework for tracing conceptual relations through language. Wherever social distinctions may have left lexical traces, masking and classification can expose how categories approach, displace or resist one another. Its contribution lies in making connections visible that are difficult to apprehend from either individual passages or aggregate counts alone: the relation between a literary departure and the linguistic expectations against which it departs, and between apparently separate categories whose histories have brought them into the same predictive field. The language model becomes, in this sense, not an automated reader but an instrument for bringing otherwise diffuse structures of linguistic association into view.

\section*{Data availability}
The code and word lists that support the findings of this study will be made openly available at https://github.com/yuxliuu89/crossing-the-gender-binary upon publication. The Gollancz SF Masterworks corpus cannot be redistributed for copyright reasons; the NovelTM derived dataset is available from the HathiTrust Research Center.

\section*{AI Disclosure Statement}
Large language models were used as part of the method of this study. Gemini 2.5 Pro (Google), DeepSeek-chat (DeepSeek) and Llama 3.3 70B (Meta) annotated the word sense of \textit{man}/\textit{men} occurrences, with the annotation checked against independent human annotation of a sample, and Gemini 2.5 Pro (Google) assigned category labels to the predictions returned by roberta-base; both procedures are described in Section 3 and note 1. Apart from the uses stated above, no generative AI was used in the creation of this manuscript. 

\section*{Acknowledgements}
The authors thank Dr Sam Halliday for his generous and attentive reading of successive drafts, and for comments that sharpened the argument at several points.

\section*{Funding}
This work was supported by Riksbankens Jubileumsfond, Change is Key! [grant number M21-0021 to H.D.]; and the China Scholarship Council under the QMUL\textendash CSC PhD Scholarship programme [grant number 202408890010 to Y.L.].

\section*{Conflicts of interest}
The authors declare no competing interests.

\section*{Notes}
\renewcommand{\notesname}{}
\theendnotes

\end{document}